\documentclass{article}
\usepackage{dcase2024,amsmath,graphicx,url,times,booktabs, tabularx}
\usepackage{siunitx}
\usepackage{enumitem}
\usepackage[utf8]{inputenc}

\DeclareSIUnit{\million}{M}
\title{The language of sound search: \\Examining User Queries in Audio Search Engines}

\name{Benno Weck$^{1,2}$,
      Frederic Font$^{2}$
      }

\address{$^1$ Huawei Technologies, Munich Research Center, Germany\\
        $^2$ Universitat Pompeu Fabra, Music Technology Group, Spain\\
        benno.weck01@estudiant.upf.edu, frederic.font@upf.edu\\
        }

\begin{document}

\ninept
\maketitle

\begin{sloppy}

\begin{abstract}
This study examines textual, user-written search queries within the context of sound search engines, encompassing various applications such as foley, sound effects, and general audio retrieval.
Current research inadequately addresses real-world user needs and behaviours in designing text-based audio retrieval systems.
To bridge this gap, we analysed search queries from two sources: a custom survey and Freesound website query logs.
The survey was designed to collect queries for an unrestricted, hypothetical sound search engine, resulting in a dataset that captures user intentions without the constraints of existing systems. This dataset is also made available for sharing with the research community.
In contrast, the Freesound query logs encompass approximately 9 million search requests, providing a comprehensive view of real-world usage patterns.
Our findings indicate that survey queries are generally longer than Freesound queries, suggesting users prefer detailed queries when not limited by system constraints.
Both datasets predominantly feature keyword-based queries, with few survey participants using full sentences.
Key factors influencing survey queries include the primary sound source, intended usage, perceived location, and the number of sound sources.
These insights are crucial for developing user-centred, effective text-based audio retrieval systems, enhancing our understanding of user behaviour in sound search contexts.
\end{abstract}


\section{Introduction}
\label{sec:intro}

Users search for foley, sound effects, and other audio elements daily, playing a crucial role in multimedia production, gaming, filmmaking, and various other creative industries. As the demand for high-quality and diverse sound assets grows, understanding user search behaviour becomes increasingly vital for developing efficient and intuitive sound search engines.
Platforms like Freesound \cite{font_freesound_2013} and FindSounds.com \cite{rice_web_2005} offer robust search functionalities to cater to this growing need for sound resources.

Unlike information retrieval involving purely textual data, multimedia retrieval --- and thus sound search --- is faced with the problem of a modality gap.
To overcome it, different forms of content-based retrieval have been proposed, such as querying by acoustic features or query-by-example \cite{kim_improving_2019}.
However, these methods are still not widely adopted and most search interfaces on the internet operate primarily with text-based search queries as input.
Despite their widespread use, there is a significant gap in research addressing how users formulate search queries on sound search platforms.
While previous studies have examined search queries for insights on semantic attributes of sounds \cite{pearce_timbral_2017}, no research, to the best of our knowledge, has systematically investigated the nature and characteristics of sound search queries, leaving a critical aspect of user behaviour unexplored.

Examining text queries is particularly valuable given the recent advancements in large language models (LLMs) \cite{minaee_large_2024}, which have significantly enhanced the feasibility of processing complex natural language inputs across various applications.
Furthermore, there is a notable trend towards multi-modal retrieval techniques, which often operate on long-form input texts \cite{zhou_state_2023,kaur_comparative_2021,levy_chatting_2023}. 
Recently a new family of audio retrieval systems focusing on cross-modal retrieval techniques have been proposed \cite{koepke_audio_2022,xie_language-based_2022,elizalde_cross_2019}.
These systems promise to retrieve audio recordings based on text queries by directly matching the text with the audio content.
This approach eliminates the need for textual metadata and potentially offers users greater expressive power.
However, real-world user needs and behaviours are often overlooked.
For example, these text-to-audio retrieval systems typically train on full-form sentence descriptions, whereas actual user inputs may not match this format.
As seen in generative systems for automatic music or image generation, user prompts tend to be short and underspecified or, more generally, be out-of-distribution in comparison to the training data \cite{chang_open_2024,xie_prompt_2023}.
This discrepancy can hinder system performance.

Prior research in web-search and information retrieval shows that people tend to search with short queries \cite{baeza-yates_modeling_2005}.
However, expectations towards systems might have shifted due to the widespread adoption of LLMs, and users might provide more text than before.
This leaves us to wonder if there is a need to investigate where on the spectrum of input length and complexity user preference falls.
This study aims to answer two questions:
\begin{description}[noitemsep]
    \item[RQ1] How would users like to search for sounds using text-only systems?
    \item[RQ2] How do people currently use text queries in a real-world sound search system?
\end{description}

In short, the contribution of our work is to shed light on user behaviour, expectations, and the status quo in sound search to guide the development of future sound search systems.
We do so by analysing actual search queries from both a custom survey and the Freesound website query logs.
Additionally, we provide insights that are important for guiding the development of user-centred, effective text-based audio retrieval systems.

\section{Method}
\label{sec:method}

In an effort to answer our research questions, we collect data from two sources: an online survey and search query logs from the sound-sharing platform Freesound. 
The survey features a mock search task designed to elicit queries that allow interpretations of user expectations towards a hypothetical sound search system backed by a limitless retrieval engine.
The query log data is selected to reflect the real-world usage of a sound search service.
We publish the data collected in the survey online.\footnote{\url{https://doi.org/10.5281/zenodo.13622537}}

\subsection{Online survey}
\label{sec:survey}
We devise a survey to collect user-written text queries with two goals in mind: i) How would users formulate queries if they do not feel restricted by the requirements of a specific search system and ii) what aspects of sounds influence the query formulation.

To ensure realistic and diverse queries, participants were assigned a search task where they could submit and potentially update their queries.
The task involved an initial stimulus in the first step and a hypothetical search result in the second step, as outlined in Figure \ref{fig:search-task}.
This setup was designed to engage participants while simulating the essential mechanics of a search engine, with the stimulus serving to define a target sound through various modalities. 
The stimuli, which were randomly assigned, were presented as either a sound recording, an image, or a text description of a sound.
In both steps, we additionally ask users what they considered important when writing or refining their query, respectively.
More specifically, they select from a list of 12 predefined aspects all that they consider relevant.
We list the aspects with a short explanation in Table \ref{tab:aspects_expl}.

\begin{figure}
    \centering
    \includegraphics[width=\columnwidth]{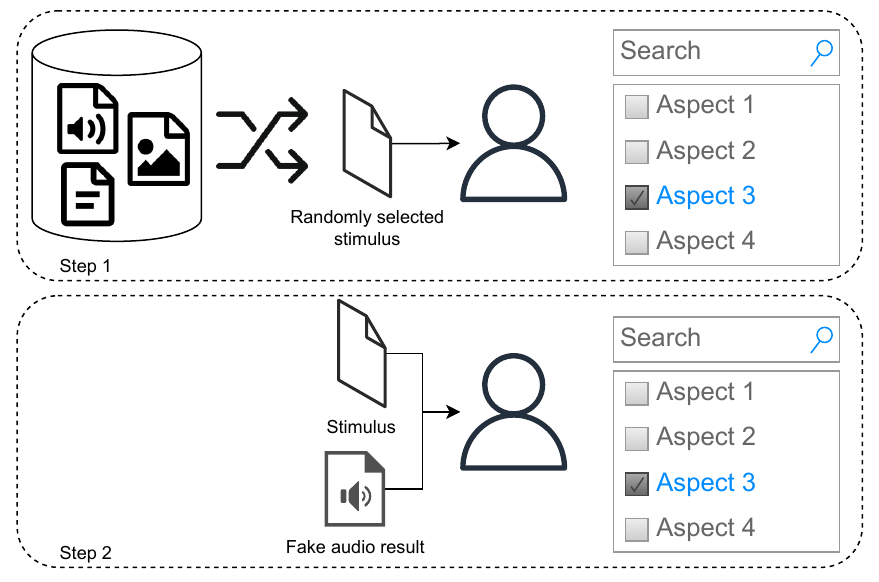}
    \caption{Schematic of the survey workflow: A participant is prompted with a randomly chosen stimulus, asked to provide a search query and indicate aspects that influenced their query. In a subsequent step, a simulated search result is presented together with the stimulus to elicit an updated query.}
    \label{fig:search-task}
\end{figure}

\begin{table}
    \small
    \centering
    \begin{tabular}{@{}lp{5.45cm}@{}}
    \toprule
        Aspect & Explanatory description \\
    \midrule
        Main sound source & The most prominent and recognisable object, entity or event in the sound. \\
        Number of sources & How many sound sources there are.\\
        Usage context & What the sound could be used for, e.g. in a movie, in a game, in a commercial\\
        Loudness & How loud or quiet the sound is.\\
        Perceived emotion & How the sound makes you feel.\\
        Recording Quality & The perceived fidelity of the sound, i.e. how clear or noisy it is.\\
        Rhythm & The perceived regularity or irregularity of the sound, e.g. repetitive/chaotic, fast/slow.\\
        Duration & How long the sound lasts, e.g. short/long.\\
        Color and/or density & The perceived quality and/or composition of the sound, e.g. bright/dark, warm/cold, harsh/smooth, simple/complex, etc.\\
        Pitch & The perceived frequency of the sound.\\
        Temporal order & The order in which events occur in time, e.g. first/last, before/after, simultaneously.\\
        Recording setting & The perceived space and environment in which the sound was recorded.\\
        \bottomrule
    \end{tabular}
    \caption{Aspect options available to survey respondents.}
    \label{tab:aspects_expl}
\end{table}

\subsubsection{Data sources}
To not be limited by the performance of an actual retrieval system and to give as much creative freedom to participants during the experiment, we do not employ any actual search engine.
Rather, we simulate retrieval results by manually mapping stimuli to audio clips prior to the experiment.
The audio clips should serve as examples of results that are somewhat relevant but not fully satisfactory and could require refinement of the query.
Moreover, the stimuli are all collected manually to relate to a wide range of potential recordings ranging from natural sounds over instrument samples to sound effects.
Specifically, we consider three different types of stimuli:

\paragraph*{Audio recordings}
The FSD50K dataset \cite{fonseca_fsd50k_2022} is chosen as a data source for our audio recording stimuli.
It features annotations for 200 sound classes and for each class, a list of example sounds is given. To obtain a stimulus-result pair, a random class is chosen and two distinct sounds of the examples are selected at random. 
\paragraph*{Images}
To acquire a set of images that can be linked to matching sounds, we first select 100 sounds from Freesound to be representative of the prevailing sound categories on the platform.
For each sound, we search for potential fitting images on the Creative Commons image platform Openverse\footnote{\url{https://openverse.org}} and select several if possible.
Through this curation process, we collect 334 image-sound pairs.
\paragraph*{Text descriptions}
We source the text descriptions from the audio captioning dataset Clotho \cite{drossos_clotho_2020}. More specifically, we synthesise summarative statements from the five crowd-sourced captions belonging to a single sound using the LLM Mixtral 8x7b \cite{jiang_mixtral_2024}.
Since the associated sound is a perfect match to the description, it can not be used as the search result. Instead, we turn to the TAU Audio-Text Graded Relevance 2023 dataset to find sounds that are relevant to the descriptions \cite{xie_crowdsourcing_2023}.

\subsubsection{Participation and Participant welfare}
To find people interested in sound search,
participants were recruited through an announcement on the Freesound website.
Participation was completely voluntarily and no compensation was given.
During the experiment, participants were free to skip a certain stimulus.
Additionally, they are offered to end their participation after completing nine search tasks and every three tasks after that.
To not bias our data through high number of annotations by individual participants, the maximum number of search tasks is 21.

Prior to participation, the survey experiment was approved by an Institutional Review Board of the Universitat Pompeu Fabra to ensure alignment with ethical guidelines and protections for human subjects in research.
The survey was fully anonymous and did not collect any personal data, safeguarding respondents' privacy and confidentiality.
Participants were informed about the objectives of the research, their tasks, and the use of their survey answers, underpinning their informed consent before contributing to the project.

\subsection{Query log analysis}
In addition to the survey, we collect anonymised system logs for search queries conducted on the Freesound website.
The text search on Freesound matches the textual metadata (user-provided sound titles and descriptions) and allows users to filter results according to various aspects including file type, sampling rate, etc.
Since our focus is on textual queries, we exclude all requests that do not specify a query or rely on search filters.
We consider search requests submitted over the course of 12 weeks from April to June 2024 and collected a total of \qty{9}{\million} queries.
Table \ref{tab:query_log} outlines the structure of the query log data.

For further data analysis, we apply a series of processing steps. 
First, all queries are case-folded.
Then, to detect search requests that were submitted by a single user in a sequential fashion, i.e. likely belonging to the same session, we group on the timestamp and anonymised IP address.
Adopting a popular baseline method in session detection, we assign requests to distinct sessions if they are separated by at least 30 minutes \cite{gayo-avello_survey_2009}.
Finally, to better understand what people are searching for, we take all queries submitted by at least 100 different IPs and manually annotate them with a single topic.  
The list of topics for annotation was adopted from the AudioSet taxonomy \cite{gemmeke_audio_2017}.
If a query term is ambiguous (e.g. `metal', `swing' or `kick') it is left unannotated.
All annotations were done by one annotator.
In total, we could annotate 978 of the 1,000 most common search queries and we share the annotations in the same repository as the survey results (see Sec. \ref{sec:survey}).

\begin{table}[th]
    \small
    \centering
    \begin{tabular}{ccc}
        \toprule
         timestamp & anonymised IP addr. & query \\ \midrule
         20240603073000 & 6ff843ba... & ``dog'' \\
         20240603073050 & 6ff843ba... & ``dog barking'' \\
         20240603073150 & d24f26cf... & ``background music'' \\
         \bottomrule
    \end{tabular}
    \caption{Excerpt of query logs collected from Freesound.}
    \label{tab:query_log}
\end{table}

\section{Results}
\label{sec:Results}

\subsection{Survey results}
In our survey, 94 participants completed a total of 706 search tasks with an average of 7.5 (median 9.0) tasks per participant.
The mean time spent on a single task is 97.8 seconds.
All three stimuli types are approximately equally represented in the data with 240 data points for image stimuli, 238 for audio, and 228 for text, respectively.
The initial query contained 4.4 tokens on average and 5.5 tokens after refinement in the second step of the search task.
Queries were slightly longer when based on a text stimulus (median: four tokens) in comparison with the other two types of stimuli (median: three tokens).
Participants chose to not update their query in the second step, or submitted the same query verbatim, in 36\% of cases.

When reviewing how participants chose to update their query, we find that most commonly the updated queries are longer by one token (32\% of the cases) or two tokens (17\%), but sometimes also keep the same number of tokens (14\%) or are shorter by at least one token (18\%). 
Examples of these updates include: `water dripping' $\rightarrow$ `slow water dripping', `car passing by' $\rightarrow$ `car passing by in distance', and `Creaking Door' $\rightarrow$ `Creaking Door Opening and Closing'.
Overall, queries consist of enumerated keywords (e.g. `drums, instrumental, live sound, music', `fishermen dock crowd'), short noun or verb phrases (`lively restaurant room', `percussion instruments', `child playing toy harmonica' ) or a combination thereof (`short clip of rapid intake of breath, moderately high pitch').
Only very few participants formed full sentences (`Man gives great speech', `Water is flowing through pipes').  
Negations are rare and mostly only present in the refined queries (e.g. `live guitar' $\rightarrow$ `live guitar no synth').

\begin{figure}
    \centering
    \includegraphics[width=1\linewidth]{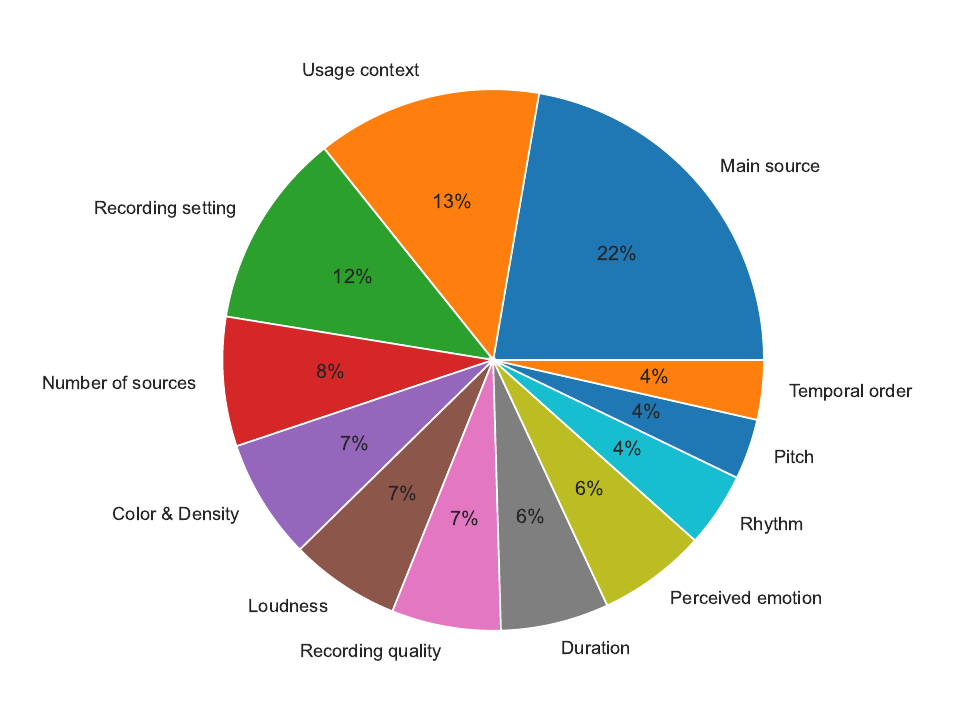}
    \caption{Distribution of aspects chosen by survey respondents to indicate what they considered important when searching for sounds.}
    \label{fig:aspects}
\end{figure}

From the Figure \ref{fig:aspects}, we can see that participants consider aspects relating to the content of the sound (main sound source, number of sources \& recording setting) most important.
Moreover, the data indicates that the usage context of a sound influences users search behaviour.
Upon closer inspection of the query terms however, we can not find this reflected in the queries, i.e. there are hardly any words that describe a usage context.
Finally, aspects related to perceptual properties (Loudness, Colour and density, Pitch) and structural attributes (Duration, Rhythm, Temporal order) of the audio recording were given less attention.

\subsection{Query log analysis}
The mean query length in the query log data is 1.8 (median: 2) and the average number of queries per session is 3.9 (median: 2).
In contrast to the survey experiment, we cannot find a significant increase in query length for subsequent requests within a session.

Figure \ref{fig:query-topics} shows the breakdown of the topics found when annotating the query log data with the first two levels of the AudioSet ontology.
We extend the ontology with a new category (`Other') including non-English texts and queries related to NSFW content. 
What stands out in this chart is that queries are generally related to a wide range of topics and span across all classes of the taxonomy.
There is a high interest in sound effects, recordings relating to objects, music, and human-made sounds.

Moreover, as a side-effect of manually annotating the topics, we identify interesting patterns in the expressions used in the queries that highlight another dimension of user search behaviour.
Table \ref{tab:jargon} lists broad categories for these expressions ranging from jargon specific to sound design, music and video production, etc. to literal use of single words to find short speech recordings.

\begin{figure}
    \centering
    \includegraphics[width=1\linewidth]{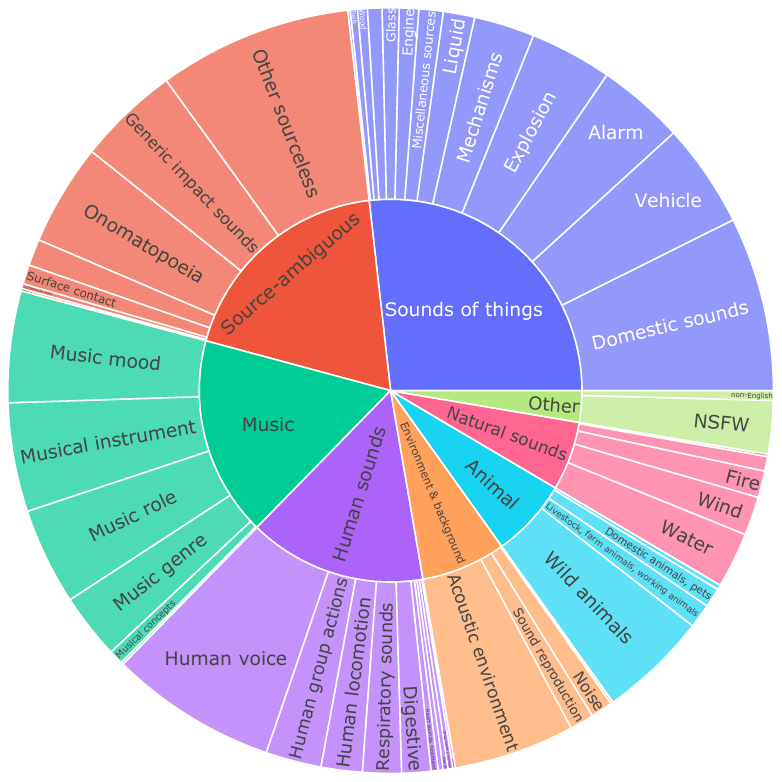}
    \caption{Distribution of topics found in the top 1,000 search queries submitted on Freesound.}
    \label{fig:query-topics}
\end{figure}

\begin{table}
    \small
    \centering
    \begin{tabular}{p{1.8cm}p{5.4cm}}
    \toprule
        Category & Search query examples \\
    \midrule
         utterances and vocables & `oh no', `yeah',  `hmm', `yay', `huh', `hello', `hey', `what' \\\addlinespace
         production jargon & `riser, `one shot', `stab', `stinger', `bumper' \\ \addlinespace
         abbreviations & `atmo', `bgm'\\ \addlinespace
         intended use & `error', `success', `correct answer', `alert', `button click', `game over', `jumpscare' \\
    \bottomrule
    \end{tabular}
    \caption{Examples of special vocabulary used in sound search.}
    \label{tab:jargon}
\end{table}

\section{Discussion and Limitations}
The above-presented results indicate that users generally tend to use short queries when using sound search systems and that there is no expectation from users that complex queries such as describing interactions between elements or temporal order are understood or helpful to achieve their search goal.
While these results are expected in light of the findings in the literature on search behaviour, the most striking difference in our study is that queries collected in the survey were significantly longer than those of the query log.
These data must be interpreted with caution since
the specifics of the Freesound search system might encourage users to submit short queries.
By default, all search terms provided in a user query must be present in a document to match the query, i.e. for a sound to be a returned result to the query ``dog barking baby crying'' all four words must be found in the metadata.
Nonetheless, it leads us to the hypothesis that users of sound search systems would provide longer and potentially more complex queries if the system supports it.

Reflecting on the latest developments in text-to-audio retrieval research, our analysis shows a discrepancy between the existing datasets and potential user input.
These datasets are usually repurposed from the task of audio captioning and we argue that they are inadequate for two main reasons.  
Firstly, datasets commonly used for evaluation and benchmarking such as the Clotho dataset might not give a reliable estimate of real-world performance due to the choice of audio recordings.
For example, the creators of Clotho purposely exclude music, sound effects, and speech recordings \cite{drossos_clotho_2020}, while it is apparent from our analysis that user interest is spread over a wide range of topics.
Secondly, the way people formulate their queries might present a challenge, as user input often takes the form of short enumerations or keywords rather than full sentences, which contrasts with the textual training and evaluation data that typically consist of complete sentences.

The generalisability of the presented results is subject to certain limitations.
For instance, one limitation of our study lies in the fact that recruitment for the experimental survey was done via the Freesound website.
Responses might be biased by participants' experience with the website's search engine.  
Furthermore, the results regarding the importance of aspects in the experimental survey provide a limited view of the participants' motivations since they are heavily influenced by the choice of stimulus. 
Finally, the relatively small sample size in topic annotations limits the comprehensiveness of our findings, highlighting the need for future research to expand and deepen our understanding of user behaviour and preferences.

Further research might also explore users' intentions in sound search since we mostly provide a view on the ``what'' dimension and not the ``why'' of search \cite{kofler_user_2017}.
We see from both sets of results that searchers (not surprisingly) focus on the main elements comprising a sound in their queries.
These results are in agreement with the observations of Giordano et al. \cite{giordano_what_2022}, who suggested that ``the most informative way to describe natural sounds verbally focuses on the properties of the sound source, rather than on sensory or acoustic attributes.''
However, understanding the underlying intentions is necessary to ultimately improve search performance satisfaction.

\section{Conclusions and Future work}
Our study analysed sound search queries from two sources: submitted by participants of an online survey and the query log of Freesound.
The results of this investigation suggest that users of sound search systems would provide longer queries if not limited by system constraints. 
The second major finding was a clear discrepancy between user-written queries and current research datasets in text-to-audio retrieval research with respect to the topics covered and the language used.
Future work should look into creating datasets specifically designed for the purpose of evaluating sound retrieval systems with user expectations and behaviour in mind.

\bibliographystyle{IEEEtran}
\bibliography{refs}

\end{sloppy}
\end{document}